\documentclass{article}
\usepackage[utf8]{inputenc}
\usepackage{times}

\usepackage{soul}
\usepackage[small]{caption}

\usepackage{appendix}
\usepackage{geometry}
\usepackage{amsmath}
\usepackage{amsfonts}
\usepackage{amsthm}
\usepackage{amstext}
\usepackage{mathrsfs}
\usepackage{amssymb}
\usepackage{graphicx}
\usepackage{subfigure}
\usepackage{algorithm,algorithmicx}
\usepackage[noend]{algpseudocode}
\usepackage{bm}
\usepackage{color}
\usepackage{natbib}
\usepackage{hyperref}
\usepackage{url}

\usepackage{float}

\usepackage{booktabs} 
\usepackage[flushleft]{threeparttable}
\usepackage{tabularx}
\usepackage{multirow}
\usepackage{xcolor}
\usepackage{hyperref}
\hypersetup{hidelinks}

\usepackage{amssymb}
\usepackage{pifont}
\newcommand{\cmark}{\ding{51}}%
\newcommand{\xmark}{\ding{55}}%

\title{A Hierarchical Destroy and Repair Approach for Solving Very Large-Scale Travelling Salesman Problem}
\author{
    Zhang-Hua Fu \textsuperscript{\rm 1,2*},
    Sipeng Sun \textsuperscript{\rm 1}, 
    Jintong Ren \textsuperscript{\rm 1,3},
    Tianshu Yu \textsuperscript{\rm 1*},    
    Haoyu Zhang \textsuperscript{\rm 1,4},\\
    Yuanyuan Liu \textsuperscript{\rm 4},
    Lingxiao Huang \textsuperscript{\rm 5},
    Xiang Yan \textsuperscript{\rm 6},    
    Pinyan Lu \textsuperscript{\rm 6,7\thanks{Corresponding author.}}\\
    {\textsuperscript{\rm 1} The Chinese University of Hong Kong (Shenzhen), China} \\
    {\textsuperscript{\rm 2} Shenzhen Institute of Artificial Intelligence and Robotics for Society,  China} \\
    {\textsuperscript{\rm 3} Bussiness School, Hohai University, China} \\
    {\textsuperscript{\rm 4} China University of Geosciences (Wuhan),  China} \\    
    {\textsuperscript{\rm 5} State Key Laboratory of Novel Software Technology, Nanjing University, China} \\
    {\textsuperscript{\rm 6} Huawei TCS Lab, China} \\
    {\textsuperscript{\rm 7} Shanghai University of Finance and Economics, China}\\
    fuzhanghua@cuhk.edu.cn, 120090026@link.cuhk.edu.cn, jintong.ren@hhu.edu.cn, \\ 
    yutianshu@cuhk.edu.cn, 
    zhanghaoyu@cuhk.edu.cn, liuyy@cug.edu.cn, \\
    huanglingxiao@nju.edu.cn,  xyansjtu@163.com, lu.pinyan@mail.shufe.edu.cn
}

\date{}

\begin{document}

\maketitle

\begin{abstract}
For prohibitively large-scale Travelling Salesman Problems (TSPs), existing algorithms face big challenges in terms of both computational efficiency and solution quality. To address this issue, we propose a hierarchical destroy-and-repair (HDR) approach, which attempts to improve an initial solution by applying a series of carefully designed destroy-and-repair operations. A key innovative concept is the hierarchical search framework, which recursively fixes partial edges and compresses the input instance into a small-scale TSP under some equivalence guarantee. This neat search framework is able to deliver highly competitive solutions within a reasonable time. Fair comparisons based on nineteen famous large-scale instances (with 10,000 to 10,000,000 cities) show that HDR is highly competitive against existing state-of-the-art TSP algorithms, in terms of both efficiency and solution quality. Notably, on two large instances with 3,162,278 and 10,000,000 cities, HDR breaks the world records (i.e., best-known results regardless of computation time), which were previously achieved by LKH and its variants, while HDR is completely independent of LKH. Finally, ablation studies are performed to certify the importance and validity of the hierarchical search framework.
\end{abstract}

\section{Introduction}

Given $n$ cities and pairwise distance $d_{ij}$ between any two cities $i$ and $j$, the travelling salesman problem (TSP) requires determining a shortest path that starts from an arbitrarily chosen city, traverses all the cities (each exactly once), and finally returns to the starting city. This problem is a famous NP-hard problem, which has many practical applications in the fields of robotic path planning, vehicle routing, etc.   

So far, TSP is one of the most studied combinatorial optimization problems. Many algorithms have been designed to solve this problem, among which the leading algorithms include the exact solver Concorde \citep{applegate2006concorde} and several heuristics, such as the LKH series \citep{helsgaun2000effective,helsgaun2009general,helsgaun2017extension} and the genetic algorithm EAX \citep{nagata2013powerful}. However, for extremely large-scale TSP instances (with millions of cities or even more), these algorithms are generally either too time-consuming, or not strong enough to obtain highly competitive solutions within a reasonable time. To our best knowledge, the largest instance solved to optimality by the exact solver Concorde is a regular instance with 85,900 cities \citep{applegate2009certification}. For larger instances, Concorde fails to terminate within a reasonable time. On the other hand, the standard version of LKH uses a 1-tree based method for pre-processing, whose complexity is about quadratic, being too high for instances with millions of cities. To overcome this drawback, for large instances, an optional choice is using POPMUSIC, i.e., Partial OPtimization Metaheuristic Under Special Intensification \citep{taillard2019popmusic}, to reduce the complexity for pre-processing. A similar difficulty arises in another powerful heuristic EAX, which performs well on small or medium sized instances, but degrades drastically due to the unaffordable computational consumption for very large instances.

In this paper, we aim to design a strong and robust heuristic with low complexity, in order to obtain high-quality solutions within reasonable time for instances with up to several million cities. For this purpose, we propose a hierarchical destroy and repair algorithm (named HDR), which is straightforward but highly competitive with respect to the leading TSP solvers, especially on large-scale instances. Overall, the main contributions are summarized from two perspectives:

\begin{itemize}
  \item \textbf{Methodologies}: Starting from a randomly generated initial solution, the algorithm applies a carefully designed destroy operator to select and delete a number of edges, then converts the original problem to a small sub-problem with bounded size and uses an effective repair operator to solve the sub-problem, and finally converts the solution to an equivalent solution with the same optimal cost. These four steps are repeated to improve the incumbent solution, until reaching a local optimum. Moreover, the key innovative component includes the hierarchical search framework, which identifies and permanently fixes a part of edges based on the historically visited local optimal solutions, and recursively compresses the original instance to a smaller instance. This framework is able to improve the robustness of the algorithm, and can be easily generalized to very large instances.
  
  \item \textbf{Results}: We carry out experiments on nineteen public benchmarks with 10,000 to 10,000,000 cities, which were also adopted by the 8th DIMACS implementation challenge. Experimental results (fairly tested on the same platform, within the same computation time) show that, our HDR algorithm is highly competitive w.r.t. the currently leading heuristics LKH3 (using POPMUSIC for pre-processing) and EAX. More interestingly, on the two largest instances respectively with 3,162,278 and 10,000,000 cities, HDR succeeds in breaking the world records (i.e., best known results regardless of computation time), which were previously both held by LKH and its variants, while HDR algorithm is completely independent of LKH. Rich ablation studies are performed to verify the importance of the hierarchical search framework.
\end{itemize}

The rest of this paper is organized as follows. After briefly introducing the related works in Sec.~\ref{sec:related}, we describe the HDR algorithm in detail in Sec.~\ref{sec:method}, then in Sec.~\ref{sec:exp} experimentally evaluate its performance with respect to several leading algorithms, and analyze the influence of the hierarchical search framework. Finally, we conclude this paper and give directions for future works in Sec.~\ref{sec:conclusion}.

\section{Related Works}
\label{sec:related}

\textbf{Traditional TSP Algorithms:} Traditional TSP algorithms (exact solvers or heuristics) are mostly designed under the context of operations research (OR). Currently, Concorde \citep{applegate2006concorde} is perhaps the best exact TSP solver, which uses linear programming as its core sub-routine, and succeeds in proving the optimality of an Euclidean TSP instance with 85,900 cities (with regular structure). However, due to the exponentially increasing solution space of the TSP, all the existing exact solvers face extreme challenges for solving larger instances with more than 100,000 cities. Notice that for random instances without regular structure, the search space can not be easily pruned, thus being much more challenging than regular instances. To our best knowledge, for random instances with more than 10,000 cities, Concorde generally fails to terminate within reasonable time. 

For the challenging instances which are difficult to be solved by exact solvers, heuristics become the main choices, since they are generally able to obtain near-optimal solutions within reasonable computation time. The leading TSP heuristics can be roughly classified into two categories, i.e., local search based heuristics and population based heuristics. The LKH heuristics \citep{helsgaun2000effective,helsgaun2009general,helsgaun2017extension} are the most famous local search based heuristics, which integrate a series of effective local search strategies based on the classic Lin-Kernighan move operator \citep{lin1973effective}, and are able to find high-quality solutions for instances with more than 100,000 cities. However, for instances with millions of cities, the standard version of LKH face big challenges due to its near quadratic-complexity pre-processing step. On the other hand, although during the early stage population based heuristics did not perform well on the TSP, but the genetic algorithm with a specifically designed edge assembly crossover operator, i.e., EAX \citep{nagata2013powerful}, shows that it is able to obtain highly competitive results on small or medium scale instances (with up to 10,000 cities). 

\textbf{Traditional TSP Algorithms Augmented by Machine Learning (ML):} Recently there are several algorithms \citep{zheng2021reinforced,zheng2021combining,xin2021neurolkh} which attempt to augment the traditional algorithms by machine learning. To do this, they first call machine learning as a pre-processing step to identify a set of candidate edges, which are directly fed into the state-of-the-art heuristics (LKH or EAX) to solve the problem. Using this mode, it is more likely to select a number of promising edges as the candidate set, thus speeding up the search for high-quality solutions. Experimental results (mainly tested on TSPLib benchmarks with up to 85,900 cities, all with known optima) show that, these hybrid algorithms yield better results than the original LKH or EAX heuristics (without ML) on a part of benchmark instances. However, it seems that most of the merits should be given to the traditional methods, while ML-based methods only slightly improve the overall performance.

\textbf{End-to-End ML-based TSP Algorithms:} In recent years, machine learning, especially deep learning techniques, develop rapidly and outperform the SOTA algorithms in many fields, such as image processing, natural language processing, etc. Motivated by these great successes, many researchers try to use ML-based methods for combinatorial optimization. Began from \citep{vinyals2015pointer}, there have been an increasing number of end-to-end machine learning based algorithms proposed to solve the TSP, including supervised learning based algorithms \citep{nowak2017note,joshi2019efficient,xing2020a}, reinforcement learning based algorithms \cite{bello2016neural,khalil2017learning,deudon2018learning,kool2018attention,Wu2020,pan2023h}, sampling based algorithms \citep{sun2023revisiting} and hybrid algorithms \citep{fu2021generalize,qiu2022dimes,sun2023difusco}. These new algorithms attempt to train neural networks based on a set of training cases, in order to capture useful information from the training data, and then transfer the knowledge to unseen cases and guide the search for high-quality solutions. 

ML for optimization is certainly a promising direction, however until now end-to-end ML-based algorithms are still not competitive enough with respect to the state-of-the-art TSP algorithms. To our best knowledge, on instances with 10,000 cities, the best result reported by end-to-end ML algorithm \citep{sun2023difusco} is 2.58\% larger than the best result reported by traditional TSP solvers. For an overall survey about ML-based TSP solvers, please refer to \citep{liu2023good,joshi2022learning,bogyrbayeva2022learning}.




\textbf{Destroy and Repair (D\&R) for Optimization:} D\&R, sometimes named as large neighborhood search (LNS), is a robust search framework for optimization. Generally speaking, any local search based TSP algorithm which tries to change a feasible TSP solution to another belongs to the D\&R family, among which POPMUSIC \citep{taillard2019popmusic} and SO \citep{cheng2023select} are two representative variants. POPMUSIC is quite straightforward, which uses a simple destroy operator (deletes a number of consecutive edges from the incumbent solution) and a simple repair operator (2-OPT based local search procedure). Although POPMUSIC is quite efficient, it alone is generally not able to obtain good enough results on large instances. Therefore, it is generally used as a pre-processing procedure to generate a set of candidate edges, which are fed into LKH to speed up the search process \citep{helsgaun2018using,taillard2019popmusic}. Motivated by POPMUSIC and the work in \citep{fu2021generalize}, in \citep{cheng2023select} the authors develop a framework that repeatedly creates a number of sub-problems, then uses a neural network to select and optimize the selected sub-problem.  

Besides the TSP, the D\&R or LNS framework has also shown strong ability for solving many other combinatorial optimization problems. For example, \citep{hottung2019neural} and \citep{li2021learning} both combine LNS with machine learning to solve vehicle routing problems (VRP), and report highly competitive results with respect to the state-of-the-art VRP algorithms. Successful applications of the LNS framework can also be found in the field of integer programming \citep{wu2021learning}, Dial and ride problem \citep{vallee2019reinsertion}, and etc. For a more comprehensive survey about LNS and its adaptive variants, please refer to \citep{mara2022survey}.


\begin{figure*}[htb]
  \centering
  \includegraphics[width=1.06\textwidth]{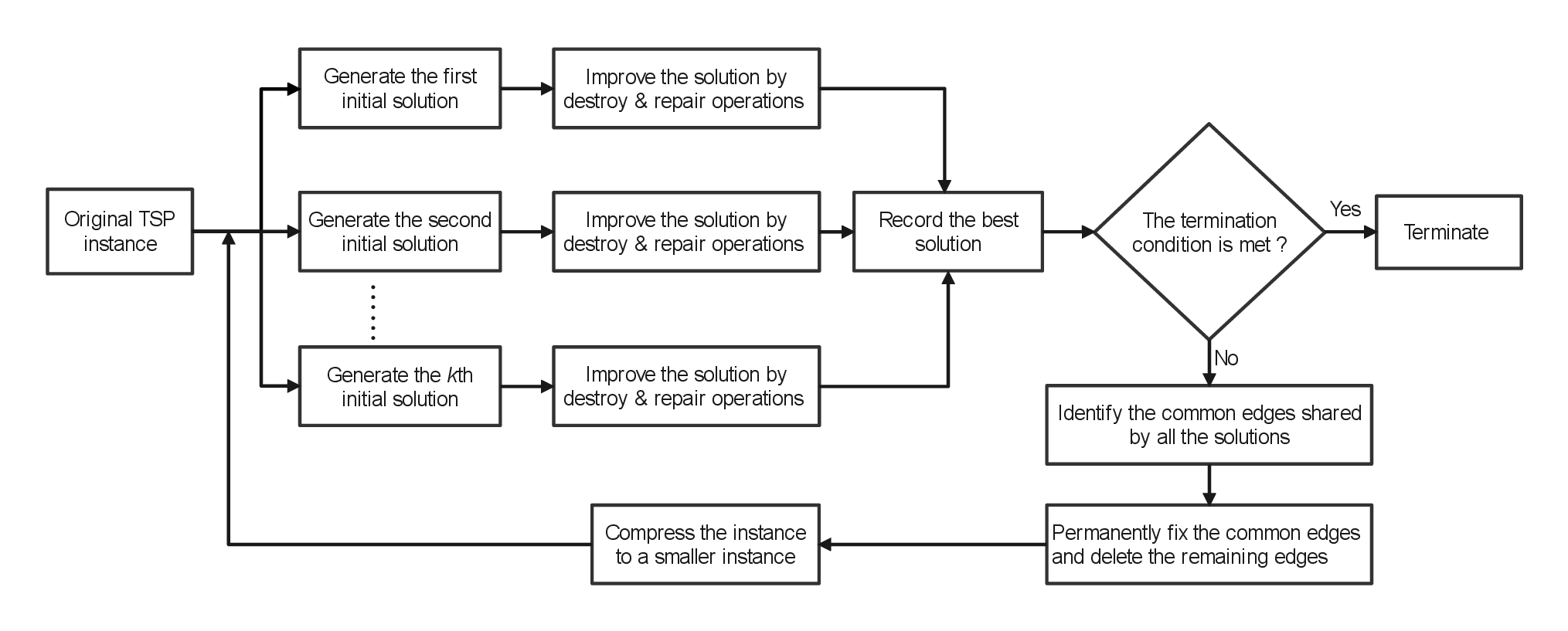}
  \caption{Pipeline of the hierarchical destroy-and-repair search framework}
  \label{Fig.framework}
\end{figure*}

\section{Methods}
\label{sec:method}

\subsection{Pipeline}

The pipeline of the hierarchical destroy-and-repair algorithm is illustrated in Fig. \ref{Fig.framework}. Given a TSP instance, the algorithm first generates an initial solution, then improves it to a local optimal solution by performing a series of destroy and repair operations. This process is repeated $k$ times, delivering $k$ local optimal solutions. After that, the algorithm identifies the common edges shared by all the $k$ local optimal solutions, then permanently fixes the common edges and deletes all the remaining edges. Thus we obtain a partial solution that generally includes a number of segments, whereas each segment consists of a series of permanently fixed edges. To reduce the problem size, the algorithm compresses each segment to only one permanently fixed edge (only retaining two endpoints), whose cost is equivalent to the sum cost of all the edges in the segment. Consequently, the original instance is compressed to a smaller instance with an equivalent optimal cost. For this new instance, the algorithm repeats the above steps, until the overall termination condition is met, then the best found solution is returned as the final solution.

More specifically, starting from a feasible initial solution, the destroy and repair process iteratively applies the following five steps to improve the solution: (1) uses a destroy operator to identify and delete a number of edges from the solution; (2) compresses the original problem (after deletion) to a smaller sub-problem; (3) uses a repair operator to obtain a near-optimal solution of the sub-problem; (4) converts the solution of the sub-problem to an equivalent feasible solution of the original problem; (5) if the new solution is better than the previous one, accepts it as the starting point of the next round of destroy-and-repair operations, otherwise abandons it and returns to the previous solution. These five steps are repeated until the number of rounds reaches the pre-set upper bound or the allowed time is elapsed, to obtain a locally optimal solution. 

\begin{figure*}[!ht]
  \centering
  \includegraphics[width=1.06\textwidth]{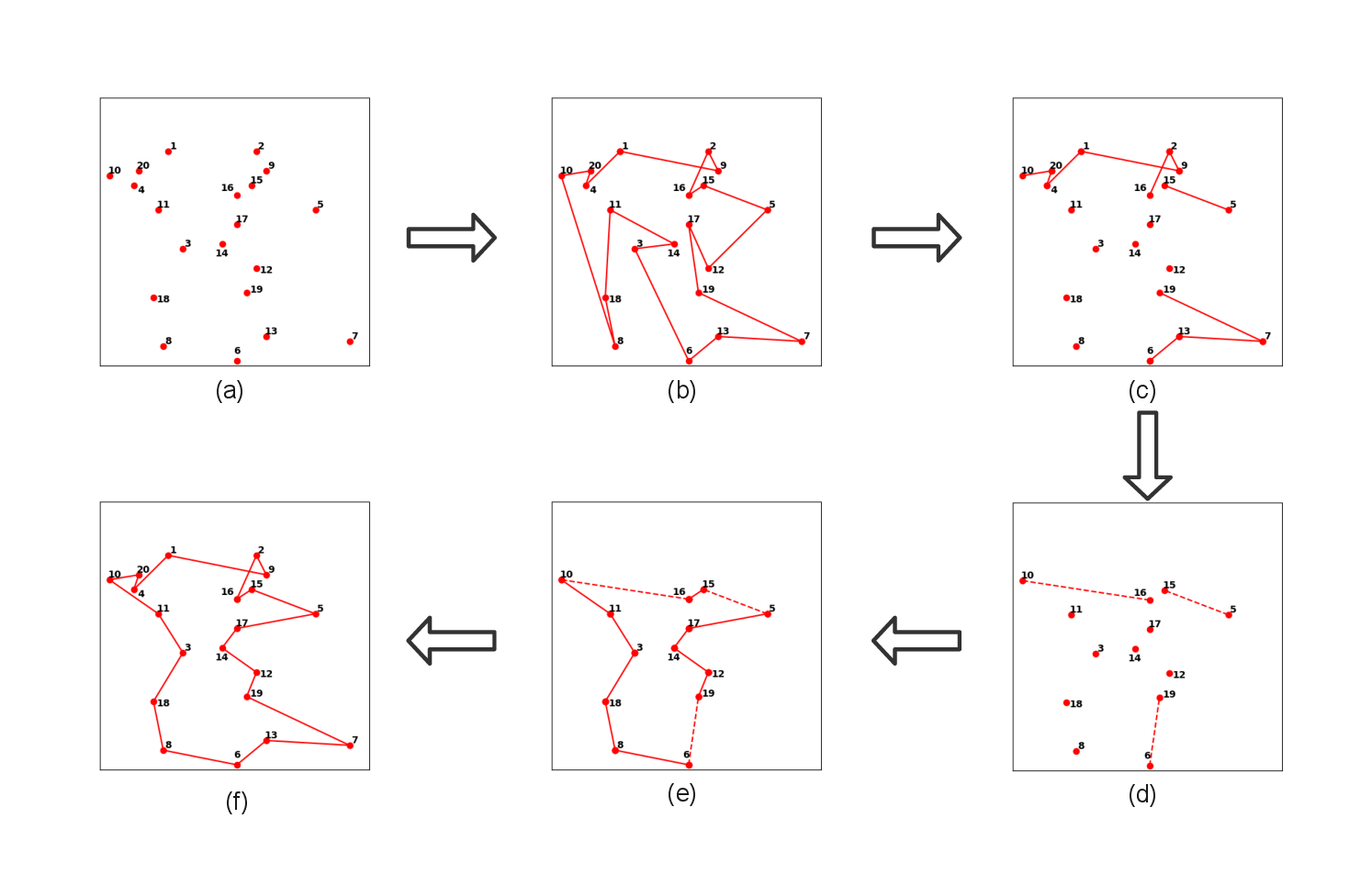}
  \caption{Improve a solution by applying a round of destroy and repair operations}
  \label{Fig.destroy}
\end{figure*}

An running example of applying the destroy and repair operators is shown in Fig. \ref{Fig.destroy}, where sub-fig. (a) is the input instance, sub-fig. (b) is a feasible solution, sub-fig. (c) is the status after deleting a number of edges, sub-fig. (d) is the new sub-problem after applying the destroy operator, sub-fig. (e) is the new solution obtained by the repair operator, and sub-fig. (f) is the equivalent solution of the original instance.  

Notice that although the search framework is quite straightforward, the basic modules of the algorithm, especially the destroy and repair operators, need to be carefully designed, since different implementations of the basic modules potentially lead to very different performance. 

\subsection{Initialization}

Throughout the whole framework, the initial solution plays an important role in the whole search process. For very large TSP instances, we can not afford a quadratic or even higher complexity for initialization. In order to generate initial solutions with low complexity, we slightly adapt the method originally developed in \citep{taillard2019popmusic}, which mainly consists of three steps: (1) identify a subset of cities as samples; (2) insert the remaining cities one by one after the closest sample city; (3) use a simple local search (2-OPT based) procedure to optimize each sub-path between several consecutive sample cities. 

As analyzed in \citep{taillard2019popmusic}, by carefully setting the number of sample cities, the complexity for initialization can be reduced from quadratic to about $O(n^{1.59})$.

\subsection{How to Destroy}

We first give several basic definitions.

\textbf{Component:} Given a feasible TSP solution $S$, a component $C$ is a set of edges belonging to $S$.

\textbf{Improvable or Non-Improvable Component:} If it is possible to improve solution $S$ by replacing component $C$ with another set of new edges not belonging to $S$, then $C$ is called as an improvable component (IC). Otherwise, $C$ is a non-improvable component (NIC).

\textbf{Minimal Improvable Component:} If component $C$ is improvable, but removing any edge from $C$ would lead to a non-improvable component, then $C$ is called an minimal improvable component (MIC).

With the above definitions, it is clear that if we want to improve solution $S$ by applying a round of destroy and repair operations, the edges deleted by the destroy operator must contain at least one MIC (otherwise no improvement is possible). Generally speaking, increasing the number of deleted edges can increase the probability of containing MICs. However, it also increases the size of the generated sub-problem, thus being more challenging for the repair operator. Therefore, the key point is how to keep a reasonable balance between the destroy and repair operations, i.e., how to identify a proper (not too large) number of edges that contain at least one MIC with as a high probability as possible.

For this purpose, we design a destroy operator which applies the following three steps to generate a smaller sub-problem based on a given feasible solution: (1) among the most rarely selected vertices (each vertex corresponding to a city), randomly select a vertex as the central vertex; (2) traverse the edges around the central vertex and deletes $m$ nearest edges (controlled by parameter $m$, excluding the permanently fixed edges which are not allowed to be deleted), to obtain a partial solution. (3) if there are some segments (each segment consists of a series of sequential edges) in the partial solution, convert each segment to only one temporarily fixed edge (only retaining two end vertices), whose cost is equivalent to the sum cost of all edges in the segment. After conducting the above three steps, the original large problem can be compressed to a smaller sub-problem with at most $2m$ vertices. 

Notice that the permanently fixed edges (most promising edges) are not allowed to be deleted during the above process. Using this method, the destroy operator mainly focuses on the less promising edges, so as to increase the opportunity of containing MICs after deleting a fixed number ($m$) of edges.   

Each sub-problem generated by the destroy operator is subsequently solved by a repair operator (described below). 

\subsection{How to Repair}

As mentioned above, the size of the sub-problem generated by the destroy operator is strictly bounded, i.e., at most twice the number of deleted edges. To solve this small sub-problem, any TSP solver can be used if it is strong and efficient enough. In this paper, we slightly adapt the EAX algorithm \citep{nagata2013powerful} as the repair operator, which is generally able to obtain close-to-optimal solutions within a few seconds for TSP instances with up to 1,000 vertices. Notice that while solving a sub-problem, the temporarily fixed edges are forcibly included in the initial solution, and are not allowed to be deleted during the search process. After solving a sub-problem, these temporarily fixed edges are all freed, being different from the permanently fixed edges which are still fixed and can never be freed. 

Based on the mapping relationship between the sub-problem and the original problem, it is easy to convert the solution of the sub-problem to an equivalent solution of the original large-scale problem.

\subsection{Hierarchical Search Approach}

We adopt a hierarchical search approach to improve the robustness and scalability of the algorithm. To do this, based on a number of local optimal solutions independently generated by the destroy-and-repair process, we identify the common edges shared by all the local optimal solutions, and permanently fix the common edges, which will never be deleted again (being different from the temporarily fixed edges which are freed once the corresponding sub-problem is solved). Besides the permanently fixed edges, all the remaining edges are deleted, to obtain a partial solution including a number of segments, whereas each segment consists of a series of permanently fixed edges. Then, from this partial solution, each segment is compressed to only one permanently fixed edge (retaining two end vertices), whose cost is equivalent to the sum cost of all the edges in the segment. After doing these steps, the original problem can be compressed into a new smaller problem, including a number of permanently fixed edges. To solve this new problem, we launch the destroy-and-repair process again to independently obtain a number of local optimal solutions, and then permanently fix a part of the edges again. This process is repeated until the maximum allowed time has been elapsed, or the new problem after compression is small enough (with less than 500 vertices, easy to be directly solved by EAX). After that, the overall best-found solution is recorded and returned as the final solution.

 
Using the hierarchical search framework, more and more promising edges are fixed recursively, thus the destroy operator can continuously expand its coverage of the area without increasing the number of deleted edges. This feature improves the opportunity of containing MICs.

\subsection{Comparisons with Existing D\&R Algorithms}
Destroy-and-repair (D\&R) is not a new search framework for the TSP, since any TSP solver which tries to change a feasible TSP solution to another feasible solution conceptually belongs to the D\&R family, among which the POPMUSIC heuristic in \citep{taillard2019popmusic} and the select and optimize strategy (SO) in \citep{cheng2023select} are two representative variants. However, HDR clearly distinguishes itself from POPMUSIC and SO as follows.

\textbf{Different Search Framework}: HDR adopts hierarchical search (key idea) to recursively compress the instance and to improve the robustness and scalability of the algorithm. No such strategy is adopted by POPMUSIC or SO.

\textbf{Different Destroy Operator}: Starting from a feasible solution (a Hamiltonian cycle), POPMUSIC simply deletes a number of consecutive edges on the cycle. SO either deletes a number of consecutive edges or deletes a number of long edges around one vertex. Being different, HDR carefully selects and deletes a number of in-consecutive edges around a central vertex. Notice that part of the edges are fixed based on historical information and are not allowed to be deleted, thus the destroy operator can pay more attention to the less promising edges, in order to increase the probability of meeting MICs. 

\textbf{Different Repair Operator}: POPMUSIC uses a simple 2-OPT heuristic as the repair operator, which yields results far away from optima even on small instances with hundreds of vertices. SO either uses a neural network, or uses LKH to solve the sub-problem. By contrast, HDR slightly adapts EAX and uses it as the repair operator, which is able to efficiently obtain near-optimal results on instances with thousands of vertices. This feature significantly enhances the search ability of our approach.

\textbf{Different Usages}: POPMUSIC is generally used for pre-processing to identify candidate edges. In SO, the destroy-and-repair operator is mainly used to escape from local optima. By contrast, HDR is executed independently to solve large-scale instances.

\subsection{Advantages of the Framework}

The hierarchical destroy-and-repair framework has advantages in four-folds:


\textbf{Simplicity}: The framework is quite straightforward and mainly consists of two modules, i.e., destroy operator and repair operator. These two modules are independent from each other, and can be easily replaced by other similar modules. 

\textbf{Low Complexity}: The algorithm mainly consists of three steps, i.e., initialization, destroy, and repair. As analyzed before, the initialization step requires a complexity of about $O(n^{1.59})$, where $n$ denotes the number of vertices. Secondly, performing the destroy operator once requires a complexity of $O(n \times log \ n)$, thus the total time elapsed by the destroy step is about $O(DR \times  n \times log \ n)$, where $DR$ denotes the number of performing destroy and repair operations, which is generally much less than $n$. Finally, for the repair step (the most time-consuming component), since the size of each sub-problem is strictly bounded within $2m$ ($m$ denotes the number of deleted edges), thus solving each sub-problem requires almost a constant time. Therefore, the total time elapsed by the repair step is nearly proportional to $DR$. Overall, the complexity of the whole algorithm is quite low, thus being easy to scale up to very large TSP instances with millions or even more vertices.

\textbf{Robustness and Scalability}: Following this search framework, to solve an instance of an arbitrarily large size, we just need to effectively solve a series of sub-problems with bounded size. This feature improves the robustness and scalability of the algorithm.

\textbf{Generalization Ability}: The destroy-and-repair approach is a general framework, which can not only be applied to solve the TSP, but also be easily generalized to other large-scale combinatorial optimization problems, such as vehicle routing, Steiner tree, task scheduling, etc (future works). 


\section{Experiments}
\label{sec:exp}


To evaluate the performance of our HDR algorithm, we program it in C++ language, then carry out experiments on a number of large-scale TSP instances. We choose the public test instances used by the 8th DIMACS Implementation Challenge\footnote{Held in 2001, see \url{http://dimacs.rutgers.edu/archive/Challenges/TSP/download.html}} as the benchmarks. Since HDR algorithm is specifically designed for solving large-scale instances, we select nineteen most challenging instances with unknown optima as benchmarks, which are classified into two categories, i.e., (1) eleven instances of the E series with 10,000 to 10,000,000 cities, whereas the cities are uniformly distributed in a 1,000,000 by 1,000,000 square, with Euclidean distances. (2) eight instances of the C series with 10,000 to 316,228 cities, whereas the cities are randomly clustered in the same square, also with Euclidean distances.

For each instance, we compare the performance of HDR with two leading algorithms LKH3\footnote{\url{http://akira.ruc.dk/~keld/research/LKH-3}.} and EAX\footnote{\url{https://github.com/nagata-yuichi/GA-EAX/tree/main/GA_EAX_1.0}.} as baselines. Notice that the standard version of LKH3 uses a 1-tree based method for pre-processing, whose complexity is about quadratic, being too high for very large-scale instances. Following the suggestion of Dr. Keld Helsgaun, we use POPMUSIC instead (by setting CANDIDATE$\_$SET$\_$TYPE = POPMUSIC and INITIAL$\_$PERIOD=1000) to reduce the complexity for pre-processing. To ensure fair comparisons, we rerun the source codes of the baselines as well as our algorithm on a uniform platform, i.e., a workstation with eighty Intel(R) Xeon(R) Gold 5218R 2.10GHz CPUs and 128GB Memory, operated by Ubuntu 18.04 system.

We do not directly compare HDR with end-to-end ML algorithms, since currently they are not competitive enough with respect to state-of-the-art TSP solvers (on instances with 10,000 cities, the gaps between the results obtained by ML algorithms and the world records are greater than 2\%, and no ML based algorithm has reported results on instances with more than 20,000 cities).


\subsection{Parameters}
\label{SubsecParameter}

HDR is quite straightforward and only relies on three hyperparameters, i.e., $m$, $k$, $l$, which respectively control the number of edges deleted by the destroy operator, the number of solutions generated during each hierarchy, and the number of destroy-and-repair operations manipulated before reaching any local optimal solution. For the sake of simplicity, we uniformly set $m$=500, $k$=10 and $l$=$\frac{n^i}{90}$, where $n^i$ denotes the number of vertices of the instance during the $i$th hierarchy. Notice that the instance is compressed after each hierarchy, thus the problem size $n^i$ reduces iteratively. 

For each test algorithm (the baselines as well as our algorithm), we independently run it ten times (executed in parallel) to solve each instance, and set the allowed time of each run to $\frac{n}{5}$ seconds for instances with less than 1,000,000 cities, or 100,000 seconds for instances with 1,000,000 or 3,162,278 cities, or 200,000 seconds for the instance with 10,000,000 cities (all including the time used for pre-processing). Notice that on the instances with 10,000 cities, LKH3 or EAX may terminate before the allowed time is elapsed (already converged).

\subsection{Results Obtained within the Same Allowed Time}

\begin{table*}[htbp] \label{Expsa1}
\fontsize{6}{7}\selectfont
\centering

\begin{threeparttable}[b]
\renewcommand{\arraystretch}{1.3}
\renewcommand\tabcolsep{1.2pt}
    \begin{tabular}{llclllllclllllclllll}
       \toprule[2pt]
        \multirow{2}{*}{Instance} & \multirow{2}{*}{World Record} && \multicolumn{5}{c}{LKH3 (using POPMUSIC for pre-processing)} && \multicolumn{5}{c}{EAX} && \multicolumn{5}{c}{HDR (Ours)} \\
        \cline{4-8}
        \cline{10-14}
        \cline{16-20}
        & && Best & Gap & Average & Gap & Time (s) && Best & Gap & Average & Gap & Time (s) && Best & Gap & Average & Gap & Time (s)\\
        \midrule[2pt]
E10k.0    & 71865826       &&71868577& 0.0038\%& 71871986.9& 0.0086\%& 16492 && \textbf{71865826}& \textbf{0.0000}\%& \textbf{71866651.9}& \textbf{0.0011}\%& 11382 && 71868057& 0.0031\%& 71871974.5& 0.0086\%& 20008\\
E10k.1    & 72031630       &&\textbf{72031630}& \textbf{0.0000}\%& 72038647.8& 0.0097\%& 16865 && \textbf{72031630}& \textbf{0.0000}\%& \textbf{72032293.5}& \textbf{0.0009}\%& 11103 && 72033351& 0.0024\%& 72037395.6& 0.0080\%& 20002\\
E10k.2    & 71822483       &&71825001& 0.0035\%& 71827952.4& 0.0076\%& 17297 && \textbf{71822483}& \textbf{0.0000}\%& \textbf{71822549.3}& \textbf{0.0001}\%& 11212 && 71824430& 0.0027\%& 71828818.4& 0.0088\%& 20009\\
E31k.0    & 127281803      &&\textbf{127292932}& \textbf{0.0087}\%& \textbf{127304022.5}& \textbf{0.0175}\%& 63247 && 127569710& 0.2262\%& 127608868.5& 0.2570\%& 63252 && 127299475& 0.0139\%& 127307516.2& 0.0202\%& 63256\\
E31k.1    & 127452384      &&\textbf{127458031}& \textbf{0.0044}\%& 127477945.5& 0.0201\%& 63251 && 127760790& 0.2420\%& 127791639.0& 0.2662\%& 63251 && 127468373& 0.0125\%& \textbf{127475923.1}& \textbf{0.0185}\%& 63249\\
E100k.0   & 225783795      &&225928098& 0.0639\%& 225952016.5& 0.0745\%& 200042 && 238006870& 5.4136\%& 238185059.1& 5.4925\%& 200034 && \textbf{225837319}& \textbf{0.0237}\%& \textbf{225844380.6}& \textbf{0.0268}\%& 200007\\
E100k.1   & 225653450      &&225788020& 0.0596\%& 225815212.6& 0.0717\%& 200046 && 237978179& 5.4618\%& 238152143.5& 5.5389\%& 200037 && \textbf{225709710}& \textbf{0.0249}\%& \textbf{225714837.8}& \textbf{0.0272}\%& 200008\\
E316k.0   & 401294850      &&401733448& 0.1093\%& 401751885.7& 0.1139\%& 632644 && 429665753& 7.0698\%& 429760312.9& 7.0934\%& 635913 && \textbf{401400270}& \textbf{0.0263}\%& \textbf{401418102.2}& \textbf{0.0307}\%& 632458\\
E1M.0     & 713176566      &&714111053& 0.1310\%& 714131779.0& 0.1339\%& 1001088 && - & - & - & - & - && \textbf{714010099}& \textbf{0.1169}\%& \textbf{714019508.7}& \textbf{0.1182}\%& 1000012\\
E3M.0     & 1267318198     &&1269276301& 0.1545\%& 1269311060.6& 0.1573\%& 1005502 && - & - & - & - & - && \textbf{1268824434}& \textbf{0.1189}\%& \textbf{1268844122.5}& \textbf{0.1204}\%& 999996\\
E10M.0    & 2253088000     &&2256905786& 0.1694\%& 2256939281.1& 0.1709\%& 2017604 && - & - & - & - & - && \textbf{2256144761}& \textbf{0.1357}\%& \textbf{2256176335.8}& \textbf{0.1371}\%& 2000658\\
\hline
Average  &    - &&   -  &  0.0644\%&   -   & 0.0714\% & - &&  -   &  2.3017\%&   -   & 2.3313\% & - &&   -  &  \textbf{0.0437\%}&   -  & \textbf{0.0477\% }& - \\

\hline
\cmark/$=$/\xmark & && 8/0/3 &&&& && 8/0/3 &&& &&  &&&& \\

    \bottomrule[2pt]
    \end{tabular}
    
    \begin{tablenotes}
    \item World Record: \url{http://webhotel4.ruc.dk/~keld/research/LKH/DIMACS_results.html} and \url{https://figshare.com/articles/journal_contribution/DIMACS_TSP_Challenge_Results/6669071} (last updated on 21.06.2018).
    \item Gap: the gap between the reported values and the world records, respectively in terms of best results and average results.
    \item Time (in seconds): accumulated time of ten independent runs, including the pre-processing time.    
    \item \cmark/$=$/\xmark: number of instances on which our HDR algorithm obtains better (“\cmark”), equal (“$=$”), or worse (“\xmark’) results with respect to the corresponding baseline algorithm.
    \end{tablenotes}
    
    \caption{Experimental results tested on the 8th DIMACS challenge E series TSP instances with unknown optima.}
    
    \end{threeparttable}
\end{table*}


\begin{table*}[htbp] \label{Expsa2}
\fontsize{6}{7}\selectfont
\centering

\begin{threeparttable}[b]
\renewcommand{\arraystretch}{1.3}
\renewcommand\tabcolsep{1.8pt}
    \begin{tabular}{llclllllclllllclllll}
       \toprule[2pt]
        \multirow{2}{*}{Instance} & \multirow{2}{*}{World Record} && \multicolumn{5}{c}{LKH3 (using POPMUSIC for pre-processing)} && \multicolumn{5}{c}{EAX} && \multicolumn{5}{c}{HDR (Ours)} \\
        \cline{4-8}
        \cline{10-14}
        \cline{16-20}
        & && Best & Gap & Average & Gap & Time (s) && Best & Gap & Average & Gap & Time (s) && Best & Gap & Average & Gap & Time (s)\\
        \midrule[2pt]
C10k.0    & 33001034       &&\textbf{33001034}& \textbf{0.0000}\%& 33030148.7& 0.0882\%& 20001 && \textbf{33001034}& \textbf{0.0000}\%& \textbf{33002600.9}& \textbf{0.0047}\%& 8986 && \textbf{33001034}& \textbf{0.0000}\%& 33005959.9& 0.0149\%& 20004\\
C10k.1    & 33186248       &&\textbf{33186248}& \textbf{0.0000}\%& 33194942.5& 0.0262\%& 20003 && \textbf{33186248}& \textbf{0.0000}\%& \textbf{33187000.8}& \textbf{0.0023}\%& 6830 && \textbf{33186248}& \textbf{0.0000}\%& 33189526.8& 0.0099\%& 17812\\
C10k.2    & 33155424       &&\textbf{33155424}& \textbf{0.0000}\%& \textbf{33156376.2}& \textbf{0.0029}\%& 20001 && \textbf{33155424}& \textbf{0.0000}\%& 33160504.4& 0.0153\%& 8812 && \textbf{33155424}& \textbf{0.0000}\%& 33157223.1& 0.0054\%& 19993\\
C31k.0    & 59545390       &&\textbf{59545390}& \textbf{0.0000}\%& \textbf{59608232.7}& \textbf{0.1055}\%& 63244 && 59631519& 0.1446\%& 59654397.3& 0.1831\%& 63251 && 59610179& 0.1088\%& 59664883.6& 0.2007\%& 63241\\
C31k.1    & 59293266       &&\textbf{59294823}& \textbf{0.0026}\%& \textbf{59334312.2}& \textbf{0.0692}\%& 63242 && 59340363& 0.0794\%& 59385325.0& 0.1553\%& 63249 && 59345936& 0.0888\%& 59371096.3& 0.1313\%& 63245\\
C100k.0   & 104617752      &&\textbf{104666387}& \textbf{0.0465}\%& \textbf{104722427.2}& \textbf{0.1001}\%& 200018 && 274393491& 162.2820\%& 280866106.0& 168.4689\%& 200018 && 104927001& 0.2956\%& 104985053.2& 0.3511\%& 199996\\
C100k.1   & 105385439      &&\textbf{105485906}& \textbf{0.0953}\%& \textbf{105540336.1}& \textbf{0.1470}\%& 200013 && 277423267& 163.2463\%& 279273610.4& 165.0021\%& 200018 && 105710390& 0.3083\%& 105802443.3& 0.3957\%& 200006\\
C316k.0   & 186834550      &&\textbf{186987077}& \textbf{0.0816}\%& \textbf{187076277.6}& \textbf{0.1294}\%& 632504 && 867906984& 364.5324\%& 871472934.9& 366.4410\%& 632559 && 187626477& 0.4239\%& 187700959.1& 0.4637\%& 632453\\
\hline
Average  &   -  &&   -  &  \textbf{0.0283\%}&   -  & \textbf{0.0836\%} & - && -   &  86.2856\%&  -   & 87.5341\% & -  &&   -  &  0.1532\%&    -   & 0.1966\% & -\\


\hline
\cmark/$=$/\xmark  & && 0/3/5 &&&& && 4/3/1 &&& &&  &&&& \\

    \bottomrule[2pt]
    \end{tabular}
    
    \begin{tablenotes}
    \item World Record: the same as in Table 1.
     \item Gap: the gap between the reported values and the world records, respectively in terms of best results and average results.
    \item Time (in seconds): accumulated time of ten independent runs, including the pre-processing time.
    \item \cmark/$=$/\xmark: number of instances on which our HDR algorithm obtains better (\cmark), equal (“$=$”), or worse (\xmark) results with respect to the corresponding baseline algorithm.
    \end{tablenotes}   

    \caption{Experimental results tested on the 8th DIMACS challenge C series TSP instances with unknown optima.}
    
    \end{threeparttable}
\end{table*}


Table 1 and Table 2 present the results obtained by HDR and the baselines. Respectively, the first two columns indicate the name of each instance (the problem type and size are included in the name) as well as the world record, i.e., the historically best-known result reported by all the existing algorithms (regardless of computation time). The following five columns give the results obtained by LKH3, including the best and average results among ten independent runs, as well as the gaps with respect to the world record, and the accumulated time of ten independent runs (including the pre-processing time). Similarly, the following ten columns respectively give the results obtained by EAX and our HDR algorithm (the best results are indicated in bold). Each row corresponds to an instance, and the second to last row gives the averaged results. Finally, the last row lists the number of instances on which HDR obtains better (\cmark), equal ($=$) or worse (\xmark) results with respect to the corresponding baseline. 

On the one hand, for the eleven instances of the E series (see Table 1), HDR clearly dominates the baselines in terms of best results and average results, especially on the large instances. Respectively, compared to LKH3, HDR obtains eight better and three worse results in terms of best results, corresponding to a lower average gap (0.0437$\%$ vs. 0.0644$\%$). Compared to EAX, although HDR obtains slightly worse results on the three smallest instances with 10,000 cities, it clearly dominates EAX on all the remaining larger instances. Notice that the performance of EAX decreases drastically with the increase in problem size. On instances with millions of cities, EAX is generally unable to find a feasible solution within the allowed time (marked as "-" in the table). By contrast, LKH3 and HDR perform much more robustly on all instances.

On the other hand, for the eight instances of the C series (see Table 2), the three competing algorithms all match the best-known results on the three instances with 10,000 cities. On the larger instances, HDR still clearly dominates EAX, and performs slightly worse than LKH3, perhaps because HDR does not fully utilize the structural information of these special instances. 

Overall, HDR is highly competitive on these large-scale instances, although there is still space for further improvement on the instances with special structures.

\subsection{Best Results Regardless of Computation Time}

\begin{table*}[htbp] \label{Expsa4}
\begin{threeparttable}[b]
\renewcommand{\arraystretch}{1.3}
    \begin{tabular}{llclllllclllllclllll}
       \toprule[2pt]
       \multicolumn{4}{c}{E Series TSP Instances} && \multicolumn{4}{c}{C Series TSP Instances}\\
       \cline{1-4}
       \cline{6-9}
       Instance & World Record & Our Result & Gap && Instance & World Record & Our Result & Gap\\
       \midrule[2pt]
E10k.0 & 71865826 & 71865826 & 0.000000\% && C10k.0 & 33001034 & 33001034 & 0.000000\% \\
E10k.1 & 72031630 & 72031630 & 0.000000\% && C10k.1 & 33186248 & 33186248 & 0.000000\% \\
E10k.2 & 71822483 & 71822483 & 0.000000\% && C10k.2 & 33155424 & 33155424 & 0.000000\% \\
E31k.0 & 127281803 & 127281803 & 0.000000\% && C31k.0 & 59545390 & 59545390 & 0.000000\% \\
E31k.1 & 127452384 & 127452384 & 0.000000\% && C31k.1 & 59293266 & 59293266 & 0.000000\% \\
E100k.0 & 225783795 & 225783795 & 0.000000\% && C100k.0 & 104617752 & 104617752 & 0.000000\% \\
E100k.1 & 225653450 & 225653450 & 0.000000\% && C100k.1 & 105385439 & 105385439 & 0.000000\% \\
E316k.0 & 401294850 & \textit{401295153} & \textit{0.000076\%} && C316k.0 & 186834550 & \textit{186836021} & \textit{0.000787\%} \\
E1M.0 & 713176566 & \textit{713176912} & \textit{0.000049\%} && & & & \\
E3M.0 & 1267318198 & \textbf{1267295637} & \textbf{-0.001780\%} && & & & \\
E10M.0 & 2253088000 & \textbf{2253040776} & \textbf{-0.002096\%} && & & & \\
\hline
Average  &  &  &  \textbf{-0.000341\%} && & & & 0.000098\% \\
       
\hline
\cmark/$=$/\xmark & & &  2/7/2 && & & & 0/7/1 \\

    \bottomrule[2pt]
    \end{tabular}
    
    \begin{tablenotes}
    \item World Record: the same as in Table 1.
    \item Our Result: the best results found by our HDR algorithm, regardless of computation time. 
    \item Gap: the gap between our results and the world records (indicated in $\mathbf{bold}$ if HDR breaks the world records).
    \end{tablenotes}    

    \caption{Best results found by HDR (regardless of computation time), compared to the world records.}
    \end{threeparttable}
\end{table*}

Furthermore, we try to verify the search ability of our HDR algorithm in case unlimited time is allowed. As mentioned above, the nineteen benchmark instances are very famous with a long history (since the 8th DIMACS competition held in 2001), which have been used as benchmarks to evaluate a large number of TSP algorithms. Currently, the world records of these instances (regardless of computation time) are all extremely difficult to break.

For each of these nineteen instances, we use HDR algorithm to search for as good results as possible (on the same platform as described above), also regardless of computation time. Table 3 lists the best results found by HDR with respect to the world records, from which we see that on all the fourteen instances with no more than 100,000 vertices, HDR succeeds in matching the world records. More interestingly, on the two largest instances of E series (respectively with 3,162,278 vertices and 10,000,000 vertices), HDR succeeds in breaking the world records respectively by 0.001780\% and 0.002096\% (indicated in bold). On the remaining three instances (two instances of E series respectively with 316,228 vertices and 1,000,000 vertices, and one instance of C series with 316,228 vertices), HDR achieves results very close to the world records, respectively corresponding to a tiny gap of 0.000076\%, 0.000049\% and 0.000787\%. These results certificate the strong search ability of HDR with respect to the state-of-the-art TSP solvers.

It is worth mentioning that HDR is completely independent from LKH, while LKH and its variants previously held the world records on all these nineteen instances. 

\newpage
\subsection{Ablation Study}
\begin{table*}[htbp] \label{Expsa3}
\fontsize{7}{9}\selectfont
\centering

\begin{threeparttable}[b]
\renewcommand{\arraystretch}{1.3}
\renewcommand\tabcolsep{2.5pt}
    \begin{tabular}{llclllllclllllll}
       \toprule[2pt]
        \multirow{2}{*}{Instance} & \multirow{2}{*}{Best-Known} && \multicolumn{5}{c}{HDR (standard version)} && \multicolumn{7}{c}{HDR-V1 (without hierarchical search)}\\
        \cline{4-8}
        \cline{10-16}
        & && Best & Gap & Average & Gap & Time (s)  && Best & Gap & Ratio & Average & Gap & Ratio &Time (s)\\
        \midrule[2pt]

E10k.0    & 71865826       &&\textbf{71868057}& \textbf{0.0031}\%& \textbf{71871974.5}& \textbf{0.0086}\%& 20008&&71880276& 0.0201\%& 6.48  &71892019.8& 0.0364\%& 4.26  &19991\\
E10k.1    & 72031630       &&\textbf{72033351}& \textbf{0.0024}\%& \textbf{72037395.6}& \textbf{0.0080}\%& 20002&&72040326& 0.0121\%& 5.05  &72054265.7& 0.0314\%& 3.93  &19993\\
E10k.2    & 71822483       &&\textbf{71824430}& \textbf{0.0027}\%& \textbf{71828818.4}& \textbf{0.0088}\%& 20009&&71839317& 0.0234\%& 8.65  &71846953.4& 0.0341\%& 3.86  &19993\\
E31k.0    & 127281803      &&\textbf{127299475}& \textbf{0.0139}\%& \textbf{127307516.2}& \textbf{0.0202}\%& 63256&&127339932& 0.0457\%& 3.29  &127349675.4& 0.0533\%& 2.64  &63232\\
E31k.1    & 127452384      &&\textbf{127468373}& \textbf{0.0125}\%& \textbf{127475923.1}& \textbf{0.0185}\%& 63249&&127507455& 0.0432\%& 3.44  &127517207.3& 0.0509\%& 2.75  &63233\\
E100k.0   & 225783795      &&\textbf{225837319}& \textbf{0.0237}\%& \textbf{225844380.6}& \textbf{0.0268}\%& 200007&&225923597& 0.0619\%& 2.61  &225931285.0& 0.0653\%& 2.43  &199993\\
E100k.1   & 225653450      &&\textbf{225709710}& \textbf{0.0249}\%& \textbf{225714837.8}& \textbf{0.0272}\%& 200008&&225797621& 0.0639\%& 2.56  &225806503.4& 0.0678\%& 2.49  &199994\\
E316k.0   & 401294850      &&\textbf{401400270}& \textbf{0.0263}\%& \textbf{401418102.2}& \textbf{0.0307}\%& 632458&&401564385& 0.0672\%& 2.56  &401575011.9& 0.0698\%& 2.27  &632444\\
C10k.0    & 33001034       &&\textbf{33001034}& \textbf{0.0000}\%& \textbf{33005959.9}& \textbf{0.0149}\%& 20004&&33009375& 0.0253\%& $+\infty$ &33039955.7& 0.1179\%& 7.90  &19993\\
C10k.1    & 33186248       &&\textbf{33186248}& \textbf{0.0000}\%& \textbf{33189526.8}& \textbf{0.0099}\%& 17812&&\textbf{33186248}& \textbf{0.0000}\%& 1.00 &33224780.0& 0.1161\%& 11.75 &19993\\
C10k.2    & 33155424       &&\textbf{33155424}& \textbf{0.0000}\%& \textbf{33157223.1}& \textbf{0.0054}\%& 19993&&33158550& 0.0094\%& $+\infty$ &33184358.3& 0.0873\%& 16.08 &19992\\
C31k.0    & 59545390       &&\textbf{59610179}& \textbf{0.1088}\%& \textbf{59664883.6}& \textbf{0.2007}\%& 63241&&59782399& 0.3980\%& 3.66  &59874513.5& 0.5527\%& 2.75  &63234\\
C31k.1    & 59293266       &&\textbf{59345936}& \textbf{0.0888}\%& \textbf{59371096.3}& \textbf{0.1313}\%& 63245&&59483955& 0.3216\%& 3.62  &59589633.2& 0.4998\%& 3.81  &63233\\
C100k.0   & 104617752      &&\textbf{104927001}& \textbf{0.2956}\%& \textbf{104985053.2}& \textbf{0.3511}\%& 199996&&105404391& 0.7519\%& 2.54  &105483692.7& 0.8277\%& 2.36  &199993\\
C100k.1   & 105385439      &&\textbf{105710390}& \textbf{0.3083}\%& \textbf{105802443.3}& \textbf{0.3957}\%& 200006&&106086256& 0.6650\%& 2.16  &106282621.3& 0.8513\%& 2.15  &199994\\
C316k.0   & 186834550      &&\textbf{187626477}& \textbf{0.4239}\%& \textbf{187700959.1}& \textbf{0.4637}\%& 632453&&188599775& 0.9448\%& 2.23  &188792091.2& 1.0477\%& 2.26  &632444\\
\hline
Average  & - && -&  \textbf{0.0834\%}& - & \textbf{0.1076\%} & - && -&  0.2158\%& 2.59 & -&  0.2819\%  & 2.62 &-\\

\hline
\cmark/$=$/\xmark   &     &&    & &   &  &  &  & 15/1/0 &   &  &  16/0/0  &   & &  \\

    \bottomrule[2pt]
    \end{tabular}
    
    \begin{tablenotes}
    \item World Record: the same as in Table 1.
    \item Gap: the gap between the reported values and the world records, respectively in terms of best results and average results.
    \item Ratio: the gap corresponding to HDR-V1 divided by the gap corresponding to the standard HDR algorithm.
    \item Time (in seconds): accumulated time of ten independent runs, including the pre-processing time.
    \item \cmark/$=$/\xmark: number of instances on which the standard HDR algorithm obtains better (“\cmark”), equal (“$=$”), or worse (“\xmark’) results with respect to HDR-V1.
    \end{tablenotes}

    \caption{Ablation studies tested on the 8th DIMACS TSP instances with less then one million cities.}
    
    \end{threeparttable}
\end{table*}


In order to certify the importance of the hierarchical search framework, based on the standard version of HDR algorithm, we implement a variant named HDR-V1 by disabling the hierarchical search strategy (i.e., no edge is permanently fixed), while all the remaining ingredients keep in accordance with the standard HDR algorithm. To compare their performances, we carry out experiments on the test instances with less than 1,000,000 cities, with the same parameters and time settings as before. Notice that for the instances with several million cities, the allowed time (100,000 or 200,000 seconds) is not enough to launch the second hierarchy, thus the hierarchical search framework is actually already disabled in the standard HDR algorithm.

Experimental results obtained by HDR and HDR-V1 are listed in Table 4 (similar to Table 1). Additionally, we add two columns to indicate the ratios, i.e., the gap corresponding to HDR-V1 divided by the gap corresponding to standard HDR. Especially, for instances on which the standard HDR algorithms succeed in matching the world records but HDR-V1 fails, the ratio is marked as $+\infty$. Obviously, a ratio greater than 1 means that HDR-V1 obtains a worse result on the corresponding instance. From Table 4, we see that after disabling the hierarchical search strategy, the performance of the algorithm decreases drastically on all the test instances. On every test instance, the gap of HDR-V1 is at least twice the gap of the standard HDR algorithm, corresponding to an averaged ratio of 2.59 and 2.62, respectively in terms of best and average results. These comparisons clearly confirm the importance of the hierarchical search framework.

\section{Conclusions and Future Works}
\label{sec:conclusion}

Hierarchical destroy and repair is a robust framework for solving large TSP instances, which is able to obtain within a reasonable time highly competitive results with respect to the state-of-the-art TSP solvers. More importantly, this approach has low complexity, thus being able to solve very-large instances with millions of cities or even more.

As a general framework, the strategies described in this paper can be easily extended to solve other large-scale combinatorial optimization problems. In the future, we plan to improve the performance on the C series TSP instances, and adapt the framework to solve other problems, such as Steiner tree problem, job shop scheduling problem, etc.  

\section{Acknowledgements}
This research was supported in part by Huawei Co. Ltd. under grant TC20210707010 and the Funding from the Shenzhen Institute of Artificial Intelligence and Robotics for Society under grant AC01202101110. Thanks to Prof. Hongyuan Zha for providing helpful suggestions. The source code would be made publicly available on MindSpore upon publication of this paper.

\bibliographystyle{named}
\bibliography{reference}


\end{document}